%% file: MIYATAKEv3.tex
\documentclass[11pt,a4paper]{article}

\DeclareMathAlphabet{\mathcal}{OMS}{cmsy}{m}{n}
\usepackage[top=25mm,bottom=35mm,left=20mm,right=20mm]{geometry} 

\usepackage{url}
\usepackage{graphicx}
\usepackage{mathptmx}     
\usepackage{mathtools}
\mathtoolsset{showonlyrefs=true}

\usepackage{color}
\usepackage{doi}
\usepackage{amsmath}
\usepackage{amsfonts}
\usepackage{amssymb}
\usepackage{amsthm}
\usepackage{bm}
\usepackage{algorithm}
\usepackage{algorithmic}

\newcommand{\ml}{\mathrm{NN}}

\newcommand{\bmx}{\bm{x}}
\newcommand{\bmy}{\bm{y}}
\newcommand{\bmh}{\bm{h}}
\newcommand{\bmf}{\bm{f}}
\newcommand{\tilh}{\tilde{\bm{h}}}
\newcommand{\esto}{\tilh_0^{-1}(\bmy_t)}
\newcommand{\estg}{\tilh_{3,0}^{-1}(\bmy_t)}

\usepackage{tikz}
\usepackage{pgfplotstable}
\usetikzlibrary{arrows,positioning,plotmarks,external,patterns,angles,
decorations.pathmorphing,backgrounds,fit,shapes,graphs,calc}
\usepackage{xcolor}
\colorlet{mygreen}{green!70!black}

\begin{document}
\title{Composing a surrogate observation operator for sequential data assimilation}

\author{Kosuke Akita\thanks{Graduate School of Information Science and Technology, Osaka University, 1-5
Yamadaoka, Suita-shi, Osaka 565-0871, Japan, akita@cas.cmc.osaka-u.ac.jp}, 
Yuto Miyatake\thanks{Cybermedia Center, Osaka University, 
			1-32 Machikaneyama, Toyonaka, Osaka 560-0043, Japan,
miyatake@cas.cmc.osaka-u.ac.jp},
Daisuke Furihata\thanks{Cybermedia Center, Osaka University, 
			1-32 Machikaneyama, Toyonaka, Osaka 560-0043, Japan,
furihata@cmc.osaka-u.ac.jp},
}
\date{}

\maketitle

\begin{abstract}
  In data assimilation, state estimation is not straightforward when the observation operator is unknown. This study proposes a method for composing a surrogate operator when the true operator is unknown.
A neural network is used to improve the surrogate model iteratively to decrease the difference between the observations and the results of the surrogate model.
A twin experiment suggests that the proposed method outperforms 
approaches that tentatively use a specific operator throughout the data assimilation process.
\end{abstract}

\section{Introduction}
Data assimilation (DA), a statistical method used in various fields, including meteorology and geology, estimates unknown states by connecting numerical simulations with observations.
DA is usually performed with a given state-space model
\begin{equation}
  \left\{
  \begin{aligned}
    \bm{x}_t & =\bm{f}_t(\bm{x}_{t-1})+\bm{v}_t, \\
    \bm{y}_t & =\bm{h}(\bm{x}_{t})+\bm{w}_t
  \end{aligned}
  \right.
\end{equation}
where $t \in \mathbb{N}=\{1,2,\dots\}$ denotes the discrete time, $\bm{x}_t \in \mathbb{R}^k$ is the state, and $\bm{y}_t \in \mathbb{R}^\ell$ is the observation.
In addition, the system and observation noise, $\{\bm{v}_t\}_{t\in\mathbb{N}}$ and $\{\bm{w}_t\}_{t\in\mathbb{N}}$, are i.i.d. Gaussian sequences of mean zero with covariances $Q_t$ and $R_t$, respectively, that is, $\bm{v}_t \sim \mathcal{N}(\bm{0},Q_t)$, $\bm{w}_t \sim \mathcal{N}(\bm{0},R_t)$.

The main task of DA is to estimate states using time-series observations.
Standard DA procedures assume that the simulation $\bm{f}_t:\mathbb{R}^{k} \rightarrow \mathbb{R}^k$ and observation operator $\bm{h}:\mathbb{R}^{k} \rightarrow \mathbb{R}^\ell$ are given in advance; however, these assumptions may be too strong for practical applications.
In this study, we are mainly concerned with the case where $\bm{f}_t$ is known, but $\bmh$ is unknown.

In the filtering step, which is part of the DA procedure, we consider the difference between the observation $\bmy_t$ and prediction of the observed variable $\bmy_t^+:=\bmh \circ \bm{f}_t \circ \bmh^{-1} (\bm{y}_{t-1})$ and obtain $\bmh^{-1} (\bmy_{t})$,
where $\bmh^{-1} (\bmy_{t})$
is a simplified notation for the DA process of estimating the state $\bmx_{t}$ from the observation $\bmy_{t}$.
The estimation of $\bmx_t$ is often denoted as $\bmx_t^{\mathrm{a}}$, $\bmx_t^+$ or $\bmx_{t|t}$ in the context of DA (see, e.g., \cite{ml1,ml2,taken1,taken2,taken3,kernel,pf,mpf}). However, we here employ the notation $\bmh^{-1}(\bmy_t)$ to emphasize its dependency on the observation $\bmy_t$ and the operator $\bmh$.
This operator $\bmh$ is often subjective, that is, it is not invertible. Thus, the notation must not be confused with the inverse map of an invertible function.

When $\bmh$ is unknown, the filtering step cannot be performed {without information about $\bm{h}$}.
Giving the surrogate operator $\tilh$ tentatively as
\begin{equation}
  \bm{y}_t \approx \tilde{\bm{h}}(\bm{x}_{t})+\bm{w}_t \label{kari}
\end{equation}
is a simple remedy, and DA can be performed with this operator; however, it does not produce appropriate estimations unless $\tilh$ well approximates $\bm{h}$.
Thus, it is important to examine the surrogate operator to ensure that the discrepancies between $\bmh$ and $\tilh$ are removed.
In this study, we propose a method for composing a surrogate operator for use in the place of the true operator.
The surrogate model is improved iteratively to decrease the 
difference between the observation $\bm{y}_{t}$ and prediction $\tilde{\bmy}_t^+:=\tilh \circ \bm{f}_t \circ \tilh^{-1} (\bm{y}_{t-1})$, where $\tilh^{-1}(\bm{y}_t)$
represents the estimation of $\bm{x}_t$ using the current surrogate operator $\tilh$.
Here, we adopt a neural network to represent $\tilh$.

In recent years, several studies on DA were conducted where only partial information on the state-space model is available.
Some studies on state estimations have proposed combining standard DA procedures and neural networks for situations when $\bm{f}_t$ is unknown (see, e.g.,~\cite{dda1,dda2,ml1,ml2}).
In addition, as another approach that does not employ neural networks, Hamilton et al.~\cite{taken1,taken2,taken3} proposed a new filter named the Kalman--Takens filter, for situations in which either $\bm{f}_t$ or $\bm{h}$ is unknown.
Berry et al.~\cite{kernel} proposed an approach that employs the kernel method when $\bmh$ is unknown, with certain restrictions.
Although our study is inspired by these studies, our approach differs from theirs in that we decided to use a neural network when $\bm{h}$ is unknown.

\section{Composing a surrogate to the true observation operator}

\subsection{Proposed method}

The key idea is to train a neural network that represents the surrogate operator $\tilh$ to minimize the discrepancy between the observation $\bm{y}_{t}$ and
prediction for the observed variable $\tilde{\bmy}_t^+$.

The outline of our proposed method is as follows:
The entire time interval $1 \leq t \leq T$ is divided into $M$ sub-intervals as
$1 \leq t \leq T_1, \ldots,  T_{M-1}+1 \leq t \leq T_M(=T)$, and an initial surrogate observation operator is provided.
In each sub-interval, we perform the DA process with the current surrogate operator and a learning process to upgrade the surrogate operator before moving to the next sub-interval.
This process is repeated until we reach the final sub-interval.

Below, we describe the procedure we use to update the surrogate operator.
The following notation is used in this paper:
Let $\tilh_0$ denote the initial surrogate operator.
For the $m$-th subinterval $T_{m-1}+1 \leq t \leq T_m$, we intend to upgrade the surrogate operator $\tilh_{m-1}$ to $\tilh_{m}$.
We represent the surrogate operator $\tilh_{m}$ as the initial surrogate operator $\tilh_0$ plus a neural network $\Delta \tilh^{\ml_m}$:
\begin{equation}
  \tilh_m :=\tilh_0 + \Delta \tilh^{\ml_m}. \label{surrogation}
\end{equation}
Here, we consider a standard fully connected network for $\Delta \tilh^{\ml_m}$ and determine the number of hidden layers and units according to the dimensions of states and observations, $k$ and $\ell$, respectively (an example of the selection is given in the next section).
Note that $\tilh_0$ is not necessarily a neural network.

The network $\Delta \tilh^{\ml_m}$ should be constructed to compensate for the discrepancy between
the true operator $\bmh$ and initial surrogate operator $\tilh_0$.
Our basic idea is to use the training data with input $\bmf_t \circ \tilh^{-1}_m(\bmy_{t-1})$ and output $\bmy_t-\tilh_0 \circ \bmf_t \circ \tilh^{-1}_m(\bmy_{t-1})$ and learn the parameter ${\bm{p}}_m$ of $\Delta \tilh^{\ml_m}$ to reduce the following cost function:
\begin{equation}
  C({\bm{p}}_m):= \frac{1}{2}\sum_{t=T_{m-1}+1}^{T_m} C_t({\bm{p}}_m),
\end{equation}
where
\begin{align}
  C_t({\bm{p}}_m):= & \Big\| \left( \bmy_t-{\tilh_0} \circ {\bm{f}_t \circ \tilh_m^{-1}(\bmy_{t-1})}\right)    - \Delta \tilh^{\ml_m}\circ {\bm{f}_t \circ \tilh_m^{-1}(\bmy_{t-1})} \Big\| ^2 \\
  =                 & \Big\| \bmy_t-{{\tilh_m}} \circ {\bm{f}_t \circ \tilh_m^{-1}(\bmy_{t-1})}\Big\|^2.
\end{align}
Note that $\tilh_m^{-1}$ is not an inverse map of $\tilh_m$ as explained in Section~1 and $\bm{f}_t \circ \tilh_m^{-1}(\bmy_{t-1})$ is obtained by a DA procedure involving the filtering step by using the surrogate operator $\tilh_m$.
We intend to find an optimal parameter $\tilde{\bm{p}}_m$ such that
\begin{equation*}
  \tilde{\bm{p}}_m = \min_{\bm{p}_m} C(\bm{p}_m).
\end{equation*}
Convergence to a global minimum may be difficult; however, as is often the case in the machine-learning context, the gradient descent method
\begin{equation}
  \bm{p}_m \leftarrow \bm{p}_m - \alpha \nabla_{\bm{p}_m}C(\bm{p}_m) \label{parariron}
\end{equation}
with the learning rate $\alpha$ or the stochastic gradient descent method is used,
and, if certain convergence criteria are met, we proceed to the next sub-interval.

However, because the iteration~\eqref{parariron} requires the estimate $\tilh_{m}^{-1}(\bmy_{t-1})$ using DA, computing the gradient $\nabla_{\bm{p}_m}C(\bm{p}_m)$ can be extremely expensive.
This prompted us to consider a relaxation to define a more practical method.
The key idea is to modify the cost function and solve the corresponding minimization problem several times in the same sub-interval.
For the $m$-th sub-interval, we aim to obtain a series of surrogate operators, such as $\tilh_{m,1}, \tilh_{m,2},\dots$, hoping that $\tilh_{m,j}$ approaches $\tilh_m$ as $j$ increases.
Let ${\bm{p}}_{m,j}$ denote the parameter of the network $\Delta \tilh^{\ml_{m,j}}$.
Consider the modified cost function
\begin{equation}
  \hat{C}({\bm{p}}_{m,j}):=  \frac{1}{2}\sum_{t=T_{m-1}+1}^{T_m} \hat{C}_t({\bm{p}}_{m,j}),
\end{equation}
where
\begin{equation}
  \hat{C}_t({\bm{p}}_{m,j}):=  \Big\| \bmy_t-{\tilh_{m,j}} \circ \bmf_t \circ \tilh_{m,j-1}^{-1}(\bmy_{t-1}) \Big\|^2.
\end{equation}
Using the training data set
\begin{align} \label{mjdata}
  \mathcal{D}_{m,j}:=&\left\{ \left(
  \bmf_t \circ \tilh_{m,j-1}^{-1}(\bmy_{t-1}),  \, 
  \bmy_t-\tilh_0 \circ \bmf_t \circ \tilh_{m,j-1}^{-1}(\bmy_{t-1})
   \right)\right\},
\end{align}
we find the optimal parameter $\tilde{\bm{p}}_{m,j}$ such that
\begin{equation}\label{paramin}
  \tilde{\bm{p}}_{m,j} = \min_{\bm{p}_{m,j}} \hat{C}(\bm{p}_{m,j}), \quad j=1,2,\dots
\end{equation}
to compose $\tilh_{m,j}$. Note that $\tilh_{m,0}$ is set to $\tilh_{m,0} = \tilh_{m-1}$. 
Then, optimizing the parameter ${\bm{p}}_{m,j}$ based on \eqref{paramin} still requires the computation of the gradient 
$\nabla_{\bm{p}_{m,j}}\hat{C}(\bm{p}_{m,j})$, but we emphasize that the new cost function 
$\hat{C}(\bm{p}_{m,j})$ does not embrace $\tilh_{m,j}^{-1}(\bmy_{t-1})$.
Thus, the cost for computing the gradient $\nabla_{\bm{p}_{m,j}}\hat{C}(\bm{p}_{m,j})$ is much reduced compared with the gradient $\nabla_{\bm{p}_{m}}{C}(\bm{p}_{m})$.

The proposed method is summarized in Algorithm~\ref{alg2}.

\begin{algorithm}[t]
  \caption{The algorithm of our proposed method}
  \label{alg2}
  \begin{algorithmic}[1]
    \REQUIRE{a surrogate observation operator $\tilde{\bm{h}}_0$}
    \STATE{Divide the entire time interval into $M$ sub-intervals}
    \begin{equation}
      1 \leq t \leq T_1, \ldots,  T_{M-1}+1 \leq t \leq T_M(=T) \notag
    \end{equation}
    \FOR{$m=1,\ldots,M$}
    \STATE{Set $\tilh_{m,0} = \tilh_{m-1}$}
    \FOR{$j=1,\ldots,J$}
    \STATE{Obtain the estimate $\tilh_{m,j-1}^{-1}(\bm{y}_t)$ by DA}
    \STATE{Generate the training data set $\mathcal{D}_{m,j}$}
    \STATE{Optimize the parameter ${\bm{p}}_{m,j}$ of $\Delta \tilh^{\ml_{m,j}}$}
    \STATE{Set $\tilh_{m,j} \leftarrow \tilh_{m,j-1}$}
    \ENDFOR
    \STATE{Set $\tilh_{m} = \tilh_{m,J}$}
    \ENDFOR
  \end{algorithmic}
\end{algorithm}

\subsection{Remarks}
\label{subsec:remark}

Note that one should not necessarily perform the inner update $\tilh_{m,j}$
until certain \emph{convergence} criteria are met.
Preliminary experiments suggest that only a few updates often produce satisfactory results. Therefore, predetermination of the number of updates $J$ is recommended.

In our proposed method, we represent the surrogate operator $\tilh_{m}$ as the initial operator $\tilh_0$ and a neural network.
Other variants are also worth considering depending on applications.
For example, one can perform a similar algorithm representing $\tilh_{m}$ with a neural network or as $\tilh_{m-1}$ plus a neural network.


\section{Numerical Experiment}
\label{sec:numexp}

As a toy problem, we perform a twin experiment using the Lorenz-96 system
\begin{equation}
  \frac{dx_i}{d\tau}=(x_{i+1}-x_{i-2})x_{i-1}-x_{i}+F,
\end{equation}
where $x_{i}$ is the $i$-th component of the state vector $\bm{x} \in \mathbb{R}^k$, $x_{-1}:=x_{k-1}$, $x_{0}:=x_{k}$, and $x_{k+1}:=x_{1}$.
We set $F=8$, and set $k$ and $\ell$, the dimension of states and observations, respectively, to $k=\ell=8$.

Consider the \emph{true} observation operator
$
  \bm{h}(\bm{x})=H\bm{x},
$
where
\begin{equation}
  H=
  \begin{pmatrix}
    c_1    & c_2    & 0      & \cdots & \cdots & \cdots & \cdots & 0      \\
    c_2    & c_1    & c_2    & \ddots &        &        &        & \vdots \\
    0      & c_2    & c_1    & c_2    & \ddots &        &        & \vdots \\
    \vdots & \ddots & \ddots & \ddots & \ddots & \ddots &        & \vdots \\
    \vdots &        & \ddots & \ddots & \ddots & \ddots & \ddots & \vdots \\
    \vdots &        &        & \ddots & c_2    & c_1    & c_2    & 0      \\
    \vdots &        &        &        & \ddots & c_2    & c_1    & c_2    \\
    0      & \cdots & \cdots & \cdots & \cdots & 0      & c_2    & c_1
  \end{pmatrix}
\end{equation}
with $c_1=1$, $c_2=0.5$.
In the following experiment, the operator $\bm{h}(\bm{x})$ is used only to generate observations.

The details for applying Algorithm~\ref{alg2} to the above settings are as follows.
First, we split the entire interval $1 \leq \tau_t \leq 350$ into the three sub-intervals $T_1=150$, $T_2=300$, and $T_3=350$, and set the number of updates $J$ to $J=2$ for the first and second sub-intervals.
Second, we employ the merging particle filter \cite{mpf}, one of the DA methods, to obtain the estimate $\tilh_{m,j-1}^{-1}(\bmy_t)$ (line 5 of Algorithm~\ref{alg2}). The merging particle filter was proposed to overcome ``degeneration'' which often appears and becomes problematic when a standard particle filter (see, e.g., \cite{pf}) is employed. Note that other methods can be incorporated.
Third, the training data set $\mathcal{D}_{m,j}$ is generated based on \eqref{mjdata}. Finally, we assume that the structure of the neural network $\Delta \tilh^{\ml_{m,j}}$ consists of four hidden layers and 80 units for each hidden layer,
and we use the hyperbolic tangent function
$
  \tanh(u)=(e^u-e^{-u})/(e^u+e^{-u})
$
as an activation function except for the connections to the output layer.

Figs.~\ref{fig:est1} and~\ref{fig:est5} compare the time-series of states for the 1st and 5th components in the final sub-interval $300 < \tau_t \leq 350$.
The results of the proposed method
$\tilh_{3,0}^{-1}(\bmy_t)$ (blue) are compared with the true states $\bmx_t^*$ (red) and estimates $\tilh_0^{-1}(\bmy_t)$ (black).
We observed that at most time points, our estimates $\tilh_{3,0}^{-1}(\bmy_t)$ more accurately describe the true states $\bmx_t^*$ than the estimates $\tilh_0^{-1}(\bmy_t)$.
Figs.~\ref{fig:err1} {and~}\ref{fig:err5} show the error between the true state and its estimation for the 1st and 5th components, respectively, in the final sub-interval.
Each green point represents
  $( |\esto_{[i]}-{\bmx_t^*}_{[i]}|, |\estg_{[i]}-{\bmx_t^*}_{[i]}|)$,
where {$[i]$ denotes the $i$-th component of the vector.}
Because we consider the final sub-interval, a total of $1000\,(=(T_3-T_2)/0.05)$ green points are plotted.
If the number of green points below the red line (defined as $n(B)$) is greater than that of the green points above the red line (defined as $n(A)$),
this indicates that the surrogate observation operator $\tilh_{3,0}$ composed by our method
is more effective than the initial surrogate operator $\tilh_0$.
The improvement is measured by
\begin{equation}
\gamma:=\frac{n(B)}{n(A)+n(B)}
\end{equation}
which we refer to as the rate of improvement.
We repeated this experiment several times under the same conditions, calculated the average of the improvement rates for each component, and present the results in Table~\ref{improve}.
These results indicate that the estimates are improved for more than half of the time points for all the components.
The results support the superiority of the proposed method over sequential DA using an initial surrogate operator.

\input{fig_timecomparison1.tex}
\input{fig_timecomparison5.tex}

\input{fig_errorcomparison1.tex}
\input{fig_errorcomparison5.tex}

\begin{table}[t]
  \vspace*{-10pt}
  \caption{Average rates of improvement $\gamma$ for each component from several numerical experiments.}
  \label{improve}
  \vspace*{3pt}
  \centering
  \begin{tabular}{ccccccccc}
    \hline \hline
                          & $x_{1}$ & $x_{2}$ & $x_{3}$ & $x_{4}$ & $x_{5}$ & $x_{6}$ & $x_{7}$ & $x_{8}$\\
    \hline
    $\gamma\times100(\%)$ & 59.4  & 63.5  & 62.4  & 62.8 & 61.1  & 61.8  & 62.8  & 58.0 \\
    \hline 
  \end{tabular}
\end{table}

\section{Discussion and Conclusion}

In general, using DA to appropriately perform state estimations is challenging in the absence of information about the true observation operator $\bmh$.
In this study, we proposed a method that composes a surrogate observation operator.
The method represents the surrogate operator as the initially given surrogate operator plus a neural network and iteratively updates the operator while advancing through the pre-divided sub-intervals. The twin experiment described in Section~\ref{sec:numexp} supports that the proposed method outperforms the approach in which the initially given surrogate operator is used throughout the DA process.

Here, we discuss the limitations of the method that must be addressed.
First, it should be noted that the performance of the proposed method depends on the properties of the \emph{true} observation operator.
The proposed method tends to prefer the true observation operator $\bmh$ to be injective, in which case it performs well; however, the performance often deteriorates if the true operator is not injective.
Because observation operators are not injective in most real-world problems, it is assumed that the injectivity of the unknown observation operator is too strong.
Thus, the proposed method would need to be upgraded such that it is able to accommodate non-injective cases.
Second, the performance of the proposed method depends on the choice of the initial surrogate operator, despite the existence of several variants of the method, as discussed in Section~\ref{subsec:remark}.
When the initial operator is selected such that it differs too much from the true operator $\bmh$, it becomes difficult to properly learn the parameter $\bm{p}_{m,j}$, which could lead to inappropriate estimates.
It is thus essential to develop a method to define an initial surrogate operator.


\end{document}

%% file: fig_errorcomparison1.tex
\begin{figure}[htbp]
    \centering

    \begin{tikzpicture}
        \tikzstyle{every node}=[]
        \begin{axis}[width=5.8cm,
                height = 5.8cm,
                xmax=15,xmin=-0.8,
                ymax=15, ymin = -0.8,
                xlabel={$\big|\esto_{[1]}-{\bmx_t^*}_{[1]}\big|$},ylabel={ $\big|\estg_{[1]}-{\bmx_t^*}_{[1]}\big|$},
                ylabel near ticks,
                grid = major,
            ]
            \addplot[only marks, fill=mygreen, mark size=0.9
            ] table {
                1.303363164	1.205883698
                0.514427774	5.377776304
                1.371620397	3.989947492
                1.13123053	1.13095699
                1.213262775	0.170869384
                0.689231543	1.27542092
                0.260148719	0.962246198
                0.639400926	0.567189275
                1.160593319	0.397213218
                1.232544835	1.040595588
                1.364171525	1.52396102
                1.807431675	2.210933406
                1.49318263	1.749586829
                1.84018947	2.259185057
                1.647365061	1.921007569
                1.608172377	1.066960724
                2.098011137	1.430071357
                1.96887148	1.146614109
                2.301116633	1.06539126
                3.158031007	1.336680937
                3.340329053	1.518787789
                3.894128798	1.19776078
                4.257915877	1.058721971
                3.730053611	0.599743123
                4.022957622	0.247421937
                4.274397776	0.35790341
                3.762005234	0.335297648
                2.172083045	0.464087079
                0.57372521	1.421325901
                3.662531506	1.927904178
                6.144594473	1.927280293
                6.50278008	0.82179509
                4.839724923	0.555075414
                2.805041488	1.171501218
                1.177757081	0.502577906
                0.765500144	0.877507959
                0.130748718	0.234863204
                1.373721963	0.265830743
                2.07732726	0.512370524
                1.615548958	0.154705918
                0.631649564	0.020545979
                0.790620303	0.176650026
                2.553727227	0.91063684
                2.133886803	0.980213879
                1.949228973	1.672454894
                1.663656647	2.660160078
                0.32721705	2.183925821
                0.408073862	2.776194306
                0.02977292	2.761475026
                1.236654718	2.092697472
                3.308072882	1.658489376
                5.515979198	1.307294088
                7.651286181	1.249583703
                10.21227871	2.064641286
                11.32825787	1.658742169
                11.08104253	1.271041431
                10.73070174	1.244202145
                9.739198831	0.706175836
                8.700998307	0.491314552
                7.457720995	0.992581409
                6.096615926	1.524310268
                4.240330404	2.292995575
                2.387905265	3.409835177
                1.299581556	4.283737789
                0.012139808	4.916173445
                1.833447299	5.141658773
                3.463578079	5.073830632
                4.12485538	3.730764182
                4.294864982	2.346840224
                3.972939246	1.276902504
                3.526806912	0.088258724
                3.598056854	0.478329454
                3.702229604	0.903773223
                4.430092536	0.606321789
                5.192546338	0.027970602
                5.218927659	0.356733034
                4.490507726	0.669670816
                4.321913141	0.324954756
                3.350625472	0.682863072
                1.507268422	1.187680285
                0.378958518	1.612265891
                0.335724744	1.962323535
                0.721958626	1.87867265
                1.184845446	2.104092445
                0.753774355	1.432152952
                0.098551665	0.386255253
                0.938904004	0.298280068
                1.168841326	0.075232552
                1.82177833	1.16947288
                1.843265772	1.167024457
                2.99357447	2.004615932
                3.141960156	1.865673634
                4.069051841	1.898430038
                3.60232596	1.822214589
                4.027682042	2.283346904
                4.018531672	2.216326379
                3.475460457	2.772388299
                3.471086042	3.687068193
                2.81869492	3.651854859
                2.038112509	3.526428779
                1.175181662	2.911827594
                1.046849969	2.566452161
                1.316158437	1.397275219
                2.775428736	0.106221429
                3.770678803	0.047395135
                4.27490846	0.540315787
                3.512089215	0.233988094
                2.3709412	0.816443912
                1.247337698	0.891522313
                1.050284661	0.138298744
                2.172921191	0.262772874
                2.955905929	0.688032489
                2.240223928	0.331576417
                0.879895642	0.090217891
                0.365613335	0.058695231
                1.97273947	1.211861367
                3.653192051	2.925568439
                4.365008142	3.759012483
                5.238797055	4.681200013
                5.060689406	4.763295068
                3.515872893	4.775253917
                0.59425687	4.048826992
                0.91946416	2.450685421
                0.528405196	0.390807696
                0.85095189	1.661808553
                2.190329263	2.116603743
                3.226241895	1.958054604
                3.691074037	1.806212582
                3.698071022	0.606546285
                3.725676243	0.099201923
                3.639911148	0.995795002
                2.385775498	0.98020705
                2.015389058	1.390031937
                1.322459007	1.406216497
                1.602063766	1.505702031
                0.766769155	1.261699176
                0.620196239	1.312809919
                1.196607065	1.24375991
                1.856197081	0.820751424
                1.270232566	1.034467168
                1.032330302	2.0286889
                1.802181052	1.89357715
                2.314209922	1.501453023
                1.685321871	1.891169032
                2.331433789	1.260640342
                2.297452947	0.964559945
                1.307725276	1.151782397
                1.418503147	0.566380331
                0.748815575	1.20376581
                0.740130432	0.664528148
                1.620643282	0.218420445
                1.657696301	0.009135041
                2.055114128	0.004447734
                3.203164378	0.287923947
                3.444208763	0.227995427
                3.843598813	0.583832954
                3.44868888	0.720665978
                3.448155497	1.194655098
                3.748020093	1.60315236
                5.058359332	1.736011016
                5.657031144	1.263990116
                5.777625927	0.731348892
                6.739028026	1.06876082
                6.8882507	1.154899328
                5.783328563	1.406840841
                2.513244904	1.674896383
                1.488879687	1.759730304
                5.089846215	1.989961139
                8.751382182	0.918196585
                10.5264723	0.755479165
                10.29875583	0.259290174
                9.97149979	0.086049694
                8.786444972	0.560555922
                7.051194143	0.147091176
                5.030121081	0.307016746
                4.044807083	0.691076187
                2.415569673	1.749139227
                1.611151638	1.970739996
                0.121318633	2.740189513
                0.837149691	2.998210873
                1.792452334	3.387533766
                2.602805132	3.854436052
                3.856604691	4.275686714
                4.596292828	5.26081278
                4.307424433	4.883875398
                3.677562341	5.230777827
                2.544578078	5.275020819
                1.342110937	4.748461147
                0.751195906	3.382739846
                2.400104342	1.866555326
                2.930403458	0.100197987
                2.698527603	1.393405569
                2.003608637	1.404078652
                1.638633767	2.267860541
                1.234288341	2.321441513
                0.563776564	1.764421116
                0.278426104	1.534713554
                0.403281821	1.515973566
                0.210081317	1.072331822
                1.559946129	0.198114018
                2.199052387	0.640694894
                0.213189256	0.495348718
                0.376937045	0.543229303
                0.223569319	1.166028816
                0.242963356	0.756211491
                0.197257855	1.684959622
                0.142185098	1.916528078
                0.687481282	0.614645134
                1.555097654	0.119491009
                2.410066075	0.468320162
                4.008052993	0.339242949
                5.624916702	0.820293233
                6.29315516	1.07808134
                6.748090114	1.066333799
                6.049429644	0.495742703
                5.729252107	0.140904209
                5.432209686	0.232241037
                3.370664588	1.855132145
                1.504794031	2.987724336
                1.968206494	3.976953202
                5.062777472	3.594413728
                7.340244588	1.472598914
                8.965827704	0.894511982
                9.106289558	1.910169478
                8.126671649	2.812411825
                6.819921821	2.22647964
                4.41374853	2.389569144
                1.836713187	3.13591164
                0.243961071	4.072959695
                0.763806773	3.947727013
                0.607490812	3.28524119
                0.552530725	3.198762241
                0.500634698	2.335783219
                1.091794845	1.441082228
                1.451926743	0.375061054
                0.97812374	0.495858264
                0.651511673	0.357387871
                1.655461067	1.317280282
                1.463161118	2.596050752
                1.454677808	2.196570222
                1.136057867	2.578840532
                1.585771773	2.380207273
                2.577008866	1.525492186
                2.488237588	1.077333039
                2.014747761	0.470339933
                1.822664482	0.068902546
                0.758641393	1.153946772
                1.55133462	1.073912596
                1.862965062	0.625846518
                1.623445488	0.84390786
                1.221563382	0.385712658
                1.357599022	0.219047596
                0.141257166	0.621565507
                0.325747863	1.733195763
                1.845791767	1.437514793
                1.775347957	1.139320772
                2.885761189	0.430895573
                3.427530807	1.828345769
                3.179911611	2.051988496
                3.439869261	2.894486379
                2.892367055	3.440124562
                1.987743761	3.303493003
                1.586486289	3.256333414
                1.619444542	3.073369071
                1.397470377	2.537693363
                1.769147414	2.271694907
                1.462625983	2.180561668
                0.741036601	1.456496705
                0.592632753	1.064876963
                0.121402102	0.296693739
                0.496074869	0.737512305
                0.395065242	0.867939656
                0.722344308	0.136661092
                1.246956522	0.686484664
                1.341546469	0.411609324
                0.953230324	0.124034853
                0.776651805	0.276342198
                0.918477299	0.604015458
                1.052687087	0.110073312
                0.499552738	0.664449409
                0.174749315	1.052135687
                0.032604029	1.344603332
                0.189468355	1.473253251
                0.740283005	1.031004957
                1.140390851	0.776357636
                0.726024832	1.205013915
                0.760708656	1.174995761
                1.050008797	1.079358004
                1.815269106	0.406782024
                2.88822358	0.307978506
                4.019704407	1.843368829
                4.198330224	2.624838177
                4.25287927	3.951987764
                3.37068382	4.607397269
                0.919737248	4.703568561
                1.68822765	1.288083712
                3.51567072	2.23267274
                3.773117779	3.844158413
                4.879312718	4.644208256
                5.35617839	5.006341256
                5.540571675	5.358052388
                4.743398123	5.328372204
                2.936813487	4.70142389
                0.909068946	3.375961369
                1.346890637	2.057666839
                2.499905893	0.552248377
                3.115136168	0.02254093
                2.470850303	0.262695115
                1.981159599	0.258391485
                1.803632078	0.046105364
                1.236465835	0.167455585
                0.581270557	0.649659693
                1.829592641	1.00793751
                3.130115236	1.166899163
                3.239622218	2.562919413
                2.547395559	2.951267238
                1.559847347	1.513298655
                0.372399045	0.715055063
                0.923915771	1.023472177
                0.914176351	1.259339633
                1.684836867	0.308569721
                1.947220811	0.285373353
                2.487853947	1.117562425
                1.880363092	1.693478777
                1.572881065	3.001334629
                0.034245789	2.071526617
                1.398026513	0.827626907
                0.754963793	0.616912474
                1.198111692	1.187928173
                0.269381398	1.452149001
                1.060884637	2.195269037
                1.460639367	3.353032573
                0.89308122	3.431051062
                0.297374254	3.547627639
                0.52860918	2.768165253
                1.203716815	1.683976309
                2.441687048	0.05725794
                3.850901523	1.794009518
                3.926203304	1.897381001
                3.009189954	1.556872097
                1.751199203	0.654610198
                1.280146025	2.219117735
                0.780617488	3.576982325
                0.452204315	4.209516103
                0.419741763	4.299957535
                0.353481979	4.308829135
                1.303858193	4.432996664
                3.010697161	3.315191343
                4.399070535	1.03108597
                4.519684555	0.815527855
                2.396487685	2.356367352
                1.648994501	1.992621269
                1.924979884	1.182968473
                2.698449552	0.243545455
                2.143003222	0.108101829
                1.19324745	0.942275848
                0.167832964	1.03256637
                0.226986629	1.709104628
                1.239692131	1.374897131
                2.078182248	1.275806984
                2.596475835	0.014376171
                4.166571613	0.586839085
                3.345288867	0.992856564
                3.613894061	1.581151021
                3.391187327	2.249807505
                2.824875985	1.214890637
                2.522843825	1.249935431
                2.18598839	0.534344085
                2.084929557	0.575273824
                3.076945427	1.124576801
                2.814967629	0.427447922
                2.510371721	0.274426736
                0.829333656	1.081841881
                1.172314013	1.658590221
                2.999698765	1.925964925
                4.685928107	2.006592382
                4.779454989	0.573377083
                5.158094406	0.253766579
                5.003751933	0.025770267
                4.368719388	0.54090062
                3.590130409	0.685134746
                3.080312772	0.639266522
                1.104803953	1.031655396
                0.372210362	1.163039971
                1.376651874	1.136110076
                1.684166314	0.649009654
                2.07789791	0.487200593
                3.248565265	1.161719587
                3.352531704	1.090422299
                2.911185834	0.371102981
                2.962814521	0.10075641
                2.844728922	0.866103192
                2.549512453	1.826185044
                2.142729756	3.029876622
                1.740318743	3.359465655
                1.495605828	3.286457932
                1.02871267	3.787579466
                0.116960702	3.693190782
                1.052475446	3.781473487
                2.086074996	3.882648323
                2.843774028	3.403231275
                2.923579248	2.570011626
                2.880348567	2.24020088
                2.957645242	2.290145732
                3.752294807	2.984423864
                4.380696394	3.261439562
                4.994255596	3.85323159
                4.813350954	3.518250366
                4.530310444	3.184733259
                4.710815058	1.570516602
                5.278159102	1.293992094
                5.538950136	1.226743939
                5.556539445	0.968284488
                6.143718321	1.167891132
                6.67041374	1.37580603
                6.097625695	1.453541307
                6.297029824	0.954991414
                6.083293857	1.585393345
                5.023018485	1.501242025
                5.086374827	2.31459373
                4.37731067	1.59723994
                3.885441587	1.249844449
                3.313283275	0.104096391
                2.433090409	1.409231128
                1.249701555	1.979605488
                0.309715687	2.836852466
                1.421173852	1.975674393
                2.114352905	0.021091396
                3.755452084	0.109732717
                3.067391984	0.837539495
                2.411152976	0.290084365
                1.615279098	0.261241144
                1.538867345	0.319813631
                1.555413831	0.047007683
                1.845248577	0.49892217
                1.519869731	0.383546289
                1.141315635	0.188314438
                1.001855191	0.340909393
                0.40834015	0.570720089
                0.140682843	0.796952409
                0.019633994	0.787996501
                0.024292277	0.483181265
                0.495233252	0.184929264
                1.222684836	0.109482555
                1.470099404	0.437537162
                1.033731673	0.79905518
                0.157236164	0.656348219
                1.910265222	0.266716015
                3.442937997	0.580312393
                3.028758606	0.490578977
                2.613359372	1.061422158
                2.126539078	1.636550518
                1.77730265	2.571758835
                1.998911864	3.132362822
                1.22754668	2.08878675
                0.939057479	0.762144534
                0.249673228	0.677857559
                1.817488748	1.551727789
                3.297253049	1.795217025
                4.061756647	2.56372584
                4.406251861	3.008103619
                3.839263149	3.44625861
                3.498663514	4.700775942
                2.914313352	5.094508712
                1.226761907	4.416438884
                0.229847942	3.89753419
                0.448638117	4.125317913
                0.440105497	3.515993085
                0.956615275	2.727337998
                0.482170619	1.482770496
                0.143050929	0.181802447
                0.099334616	1.573445112
                1.500335477	0.89037693
                2.399137136	1.10143508
                2.573101804	0.261616541
                2.751638053	0.381307529
                3.234903512	0.174814554
                3.010398982	0.275737678
                2.223563417	0.588794842
                1.532509761	1.545497422
                1.690642617	0.462005269
                0.962384497	0.468906157
                1.305588159	0.599796498
                1.356663227	0.539346436
                0.744393767	0.379832184
                0.206272809	0.087717726
                0.47873962	0.820911953
                0.865490049	1.163259622
                0.547150053	0.554070039
                0.924786682	0.510104899
                1.253535012	0.811644644
                1.381048455	0.782766571
                0.877277605	0.769942876
                0.630972761	0.31151896
                0.420550555	0.312984607
                0.901055627	0.54061401
                1.029819174	0.846984646
                1.549327942	0.942653837
                1.100113343	0.684790169
                0.322230246	0.718541563
                0.617241526	0.058518114
                0.898280468	0.438524055
                2.596475639	2.185758183
                4.276284326	3.487292654
                6.045253429	4.229499016
                7.216168222	4.115116138
                7.65902885	3.521997557
                7.588890353	2.620484838
                6.797939749	1.737296254
                5.252044722	1.005665187
                2.246707765	0.339012099
                0.499332554	1.337838545
                1.232421908	1.352309163
                2.569440766	1.457388016
                3.401987998	1.290999126
                3.801677846	1.207007998
                4.973863944	1.859278411
                5.534475234	1.7267637
                4.754255162	0.594406417
                4.226998471	0.7375382
                3.075922673	2.246114858
                0.774614249	3.694358118
                0.574859428	3.792298348
                2.353784581	3.229352426
                3.159944192	3.598497888
                4.442154915	0.98222849
                5.442940175	0.134800036
                4.110219901	0.989227433
                1.156587485	1.579208647
                2.073773447	3.465843429
                4.799144863	2.34370513
                5.682214739	1.257919437
                3.346014898	0.204526409
                0.893947281	1.160141318
                4.011603578	1.413144472
                5.975664072	1.336970052
                7.721889104	0.481081216
                8.009203763	0.519823382
                5.840181405	1.32351165
                4.432732449	0.770305349
                2.699106378	1.064773932
                1.350168321	0.34056922
                0.547889345	0.125478445
                1.326548195	0.533476569
                2.5203504	1.191568374
                2.716044599	0.381609333
                2.82522593	1.197754351
                1.035215538	2.10329856
                0.233026539	1.977520247
                0.475068647	0.822435123
                0.374616975	0.251654977
                1.798263641	0.850147038
                3.617315815	0.080577478
                5.086152384	1.126953189
                4.290028834	0.372812189
                5.224200364	0.330856188
                5.746535582	0.201134319
                5.657277033	0.317713931
                5.288264963	0.300487967
                5.87084901	0.045828416
                6.075908986	0.642847498
                5.739748724	1.03422881
                6.274433719	0.567890675
                7.075919942	2.216566712
                7.188377697	1.542517732
                6.368294219	1.363667112
                5.549390694	1.999960234
                3.157266839	1.893458334
                0.975409221	1.710902608
                1.856396712	1.561020628
                3.792877183	1.788853812
                5.9532026	1.221661962
                6.839588092	1.279420803
                7.513508738	0.652119838
                5.831536866	1.215873393
                5.58364322	0.516793758
                5.402035429	0.577487391
                5.467892883	0.204171162
                5.564196175	0.38467605
                5.486665041	0.660797406
                5.499279438	1.114154099
                5.666594122	0.529063624
                5.174940997	0.314776867
                3.890736545	0.457648006
                2.712436024	0.801893336
                0.980766313	1.084341255
                0.167980174	2.037960038
                0.622848068	2.297427691
                1.040273099	1.271778689
                1.41384782	0.811652754
                0.068903839	2.983901827
                1.174054487	4.995879256
                2.276885419	5.012538043
                2.621748347	3.449572192
                1.603381803	2.080341263
                0.893655297	0.149422932
                0.929696875	2.757806939
                0.331950602	4.56449755
                0.555063212	5.465037555
                0.814152818	6.12384241
                2.050188616	5.385457779
                2.680613316	4.409253208
                4.106352229	3.839912962
                4.365765832	3.467013859
                4.109699028	3.358071201
                3.421481245	3.167329783
                2.412696671	2.725580788
                0.848441631	2.002392965
                0.504130928	1.072333838
                1.837457945	0.921984251
                2.652160595	0.560143335
                2.096408624	0.807344384
                1.469471768	1.631866376
                0.269887421	1.93563192
                0.234519363	2.2172041
                0.012855	2.347565479
                0.136926822	2.532286248
                0.665498835	2.919955652
                0.417425282	2.728671672
                0.281969524	2.375412064
                1.590668943	1.462122464
                2.224319951	0.537906469
                3.084713003	0.212475793
                3.372423386	0.275094105
                3.736942025	0.396929794
                2.994466228	0.037343856
                1.859691228	0.644768433
                0.343439391	0.683903377
                0.060508604	0.789889852
                0.931224155	1.200916374
                0.564179036	0.672269057
                1.60922572	1.113197225
                1.46410831	1.037932591
                1.24671511	1.12049074
                1.468500323	1.577604382
                1.310144894	1.342275728
                1.563671797	1.353786682
                1.963923367	1.332906474
                2.90652918	0.487815426
                5.645904879	0.23281915
                6.635690753	0.087566416
                5.328946076	0.871828148
                2.250440311	1.059296111
                0.913628927	0.202489611
                2.167429541	0.436467536
                2.119408493	0.426863425
                1.60065644	1.057300367
                0.653753334	1.304052836
                0.767694177	1.538893031
                0.019847091	1.940249934
                0.965916204	1.980642354
                1.263366972	1.957988948
                1.485736396	2.580177066
                0.876964715	1.528585
                0.044831753	0.826542532
                1.795588014	0.191664772
                2.622755618	0.215514971
                3.166210086	0.301428026
                3.456184672	0.373149163
                2.770972432	0.139598955
                2.137528075	0.004345412
                1.222562842	0.239816665
                0.107265492	0.420033877
                0.839737349	0.028457827
                0.897786982	0.01171952
                0.851693284	0.186312765
                0.785599263	0.498209435
                1.966710682	0.129895868
                2.104034969	0.281590928
                1.722863619	0.433729707
                1.142858667	0.943885329
                1.306465263	0.252968935
                1.62146579	0.777625738
                1.177715586	0.125530925
                0.54061513	0.581284944
                0.421129438	0.626473801
                0.267959059	0.804428912
                0.285419912	0.796101954
                0.188764172	1.621810922
                0.700567711	1.944053012
                1.39273978	2.177578352
                0.772892644	1.447313716
                1.035318752	1.615891303
                1.12958781	0.961257967
                1.508801972	0.792622147
                1.193057995	0.786743002
                2.077353983	0.44971821
                2.416377912	1.091343937
                1.482377034	0.305157866
                0.704474656	0.686634791
                0.268085619	0.988051671
                0.846604162	1.138369267
                0.972071533	1.631647162
                0.00784402	1.99187072
                0.375566775	1.63000877
                0.402855552	1.307556252
                1.110910086	1.236319686
                1.317120078	0.895093738
                0.338474007	0.350225625
                0.366036311	0.292429332
                0.20517506	0.179381051
                0.539731919	0.381084676
                0.66108423	0.298767614
                0.326389421	1.021204027
                0.070485272	1.564856109
                0.471461448	1.462153327
                0.887232013	0.692601169
                0.758350168	0.357314806
                0.398572232	0.166060264
                0.177442142	0.248509321
                0.04199169	0.309273491
                0.443607548	0.569840603
                0.202521919	0.790387188
                0.047510282	0.876301027
                0.141960726	1.164602693
                0.656503075	1.567011618
                0.408768919	1.030654531
                0.60391283	0.472719802
                1.303548698	0.410904805
                2.026824893	0.198114496
                2.123133877	0.239876373
                2.714100373	0.269077574
                3.585361988	0.748820321
                3.051115348	0.415716836
                2.475645814	0.185402796
                1.171852917	0.373240613
                1.577471942	0.28738929
                3.962062901	0.214050731
                6.370350522	0.128517647
                4.706761039	0.213794703
                2.347122189	1.300157333
                0.530907524	1.921133233
                0.06013419	2.393491634
                1.074949442	1.310654349
                1.378878729	0.471220456
                0.943421775	0.239144163
                0.974489835	2.02442353
                0.171766752	3.357804995
                0.845929748	5.00782149
                1.15585095	6.188040748
                0.909953009	5.97658018
                0.558893238	4.794737285
                1.260518717	4.096001521
                1.950172646	3.127024676
                3.152697782	3.594508194
                4.276395349	3.640051103
                4.445246337	3.826544886
                4.152117183	4.149410448
                2.23416313	2.860532994
                0.322367039	1.347651763
                0.409823133	1.314567388
                0.20106179	1.532944249
                0.933010198	1.886797925
                1.70823983	1.382483504
                2.378171069	1.05161533
                2.407606929	1.174045177
                1.825584827	1.250473626
                1.881862919	2.20729143
                1.929412127	2.340487499
                1.636891084	2.432697969
                1.593143343	2.924212783
                1.427245868	3.0675006
                0.952640832	2.354143352
                0.015503868	0.901329023
                1.180277323	1.002075614
                2.951925328	3.149399022
                4.844618984	5.550240935
                5.103721543	5.514167104
                4.169993056	4.597050525
                3.517037675	4.38748463
                2.908456418	4.056987646
                2.241292382	2.840616378
                1.90239442	2.049418915
                1.593771662	1.003394
                1.530020186	0.821607188
                1.116950164	1.10123462
                1.164022533	0.621804248
                0.862228352	0.489973033
                1.129503458	0.357901026
                0.094362882	0.692763341
                0.643862986	0.36597506
                0.392855479	1.07861536
                0.668029601	1.283789419
                0.250454311	0.961778035
                0.161351665	0.915763189
                0.370877216	0.522335178
                1.135872432	0.05483667
                1.618952487	0.507234674
                1.413283416	0.622719019
                1.613604255	1.090515741
                2.119624181	1.66653004
                2.197152273	1.772871722
                1.385746697	1.290206206
                1.35114622	2.012120916
                1.415134276	2.077807061
                1.235089589	1.694529557
                1.795281273	1.762176189
                1.77935375	1.730122296
                1.827185925	2.183231981
                1.880533872	1.876517231
                1.886699242	1.761726802
                1.986845001	1.883634606
                2.518688646	2.860774383
                2.322353049	3.111666561
                1.338557518	1.659773482
                0.659329565	0.353229499
                0.34228415	0.473321009
                1.256722008	2.356349157
                0.079630968	2.288037507
                0.767933137	1.246501924
                1.596900016	0.95117326
                2.44614023	0.321603558
                1.824159111	0.060080197
                0.225106628	0.767438019
                2.500763697	1.346884128
                3.330173366	1.572576524
                2.794113726	1.914729583
                1.051723871	2.219603294
                1.003472222	2.454690388
                1.065765166	1.92370622
                0.591459739	1.899463579
                0.418609169	1.537986664
                0.625005941	1.218431411
                0.956696826	1.012769089
                2.15330895	0.521579099
                2.768763085	1.20620776
                3.632174359	2.073593492
                3.503460749	2.742712067
                4.23892171	4.111766239
                4.499925209	5.615494159
                3.055846798	5.643752479
                1.87377162	5.033838216
                1.257440761	3.894132159
                0.129683605	2.327276166
                0.41899264	0.525392858
                0.449626775	1.292550029
                0.544265506	2.838873136
                0.271512889	3.042039018
                0.478782558	1.583466394
                0.417835078	0.010800804
                0.207590517	0.73898516
                1.905681544	0.031996689
                3.329107454	1.731041441
                4.280067159	2.828345436
                4.368591021	3.588846795
                3.578340474	4.186955832
                3.070369462	4.058305874
                2.843986394	2.917432544
                2.378792339	2.090269381
                2.996621202	0.447044412
                2.929755089	0.585713983
                2.675276002	1.040146265
                2.371021541	1.055076995
                2.678389621	1.705836255
                1.511687361	0.801664845
                1.006971319	0.75509703
                0.703528346	0.762971286
                0.590189588	1.987850219
                0.324127861	0.57649859
                0.856626948	0.613526104
                0.699763411	0.70994427
                2.162153974	0.912370614
                2.215631102	2.009511705
                2.467947977	2.903534853
                2.754033466	4.314931855
                1.499947694	5.385956305
                0.075548736	5.262199197
                0.14390147	4.897643907
                0.941351166	4.056868354
                2.362788199	3.109559211
                4.196967708	2.288114759
                5.873069434	1.67148611
                5.823851497	1.261718763
                4.772891636	1.259465079
                4.081101707	2.136391485
                3.239161009	2.739154232
                3.397510546	3.535891058
                3.459227497	3.815755747
                3.225697966	3.75435492
                3.04382939	3.573460881
                3.450836232	3.541736634
                4.361530551	4.355671384
                5.103006105	4.046285828
                5.93081311	3.933505223
                6.858033251	3.518441721
                7.934779625	4.924886168
                8.210316361	5.947178156
                7.851407071	6.741952868
                6.945663672	6.456282855
                5.727419982	5.885602472
                4.734744943	5.436212015
                3.328305481	4.328898697
                1.772740081	2.759430932
                0.477751453	0.935476027
                0.380444041	0.094779076
                0.475711722	0.956173678
                0.323956815	0.460605634
                0.253996342	0.532492637
                0.727573108	0.080424539
                3.191693132	1.003109501
                4.122417329	1.250394382
                5.132313825	1.398575094
                6.272018203	1.882460261
                5.309212875	0.729840971
                2.827173404	0.337709227
                1.747133512	0.026159147
                4.093521567	1.567915394
                6.389353111	1.547910777
                7.214480802	1.371121646
                7.36718449	0.285383098
                6.66555121	1.06155165
                5.24249714	1.069301993
                4.26659187	1.11668884
                3.345093968	1.326751684
                1.825877715	0.30527374
                0.766472393	0.622625481
                0.197908616	1.198684403
                0.226271851	1.122598534
                0.099859531	0.74081927
                0.21801145	0.889302675
                0.34186558	0.737130486
                0.564441672	0.260215694
                0.061607457	0.185877917
                0.282222121	0.724970529
                0.634849432	1.156101088
                0.181578817	0.984216431
                0.144150307	1.391733286
                0.119782673	1.361076565
                0.522460886	1.273151183
                0.244236526	0.969740703
                0.128913459	0.718349235
                0.074825401	0.478018186
                0.133960643	0.327929731
                1.217328748	1.225993428
                2.26856589	1.040100457
                2.653012999	0.494244926
                2.362194169	0.019147089
                1.406561659	0.801895519
                1.211821757	0.095378781
                0.727669135	0.397085595
                1.047606946	0.214209304
                0.940455471	0.662140498
                0.168648646	0.297979546
                0.912888475	0.159400662
                2.310690144	1.074038359
                2.921056399	0.883738724
                3.641693752	1.144708511
                3.251766886	0.809023002
                3.709060221	1.027063509
                3.215633101	0.824601301
                2.727482883	0.486904612
                2.242361066	0.071539483
                1.677235479	0.060671087
                1.316188642	0.89710605
                1.153985511	0.929831489
                1.409965283	0.873423516
                1.530151057	0.820858799
                2.42787268	0.686583693
                2.036898426	0.514387762
                2.921415572	1.536677777
                3.742943999	2.184764146
                4.676239735	1.80824132
                4.720125148	1.328575331
                4.57319598	0.120879675
                4.927792669	0.200176959
                5.060462017	0.678278998
                5.427068705	0.296940775
                6.007025837	0.346579393
                6.267718735	0.102117167
                7.702845689	0.393562738
                9.796089563	1.464241086
                12.81943767	3.258317864
                13.39551398	3.991130106
                13.73204643	4.302846177
                12.05633326	3.947228306
                8.687303151	2.322378076
                4.526721092	1.150866284
                0.584744376	0.456167427
                2.384856511	0.220373293
                4.46035823	0.856939771
                5.129228125	2.189556419
                6.189940432	2.033404947
                4.919997474	1.634435995
                3.660048208	0.426615549
                2.426420317	0.046966311
                2.454488221	0.10917168
                3.48088013	0.111301203
                3.501796029	0.197903501
                3.189061154	0.78476581
                0.999977321	0.33293008
                0.51420234	0.923077968
                1.818984927	1.976448494
                3.720582282	3.498106709
                4.265990472	2.536597273
                5.080196144	1.105700955
                5.849431911	0.425259097
                5.828833433	1.16166212
                6.18382495	1.613270801
                5.257606701	1.61603723
                2.945039796	1.900352303
                2.150949576	2.110972211
                };
                \addplot[color=red,thick] table {
                -0.8 -0.8
                15 15
                };
        \end{axis}
    \end{tikzpicture}
    \vspace*{-9pt}
    \caption{
        The comparison of error between $\big|\esto_{[1]}-{\bmx_t^*}_{[1]}\big|$ 
        and $\big|\estg_{[1]}-{\bmx_t^*}_{[1]}\big|$ for $300<\tau_t \leq 350$.
    }
    \label{fig:err1}
\end{figure}

%% file: fig_errorcomparison5.tex
\begin{figure}[htbp]
    \centering

    \begin{tikzpicture}
        \tikzstyle{every node}=[]
        \begin{axis}[width=5.8cm,
                height = 5.8cm,
                xmax=15,xmin=-0.8,
                ymax=15, ymin = -0.8,
                xlabel={$\big|\esto_{[5]}-{\bmx_t^*}_{[5]}\big|$},ylabel={$\big|\estg_{[5]}-{\bmx_t^*}_{[5]}\big|$},
                ylabel near ticks,
                grid = major,
            ]
            \addplot[only marks, fill=mygreen, mark size=0.9
            ] table {
                    0.886766591	0.385742993
                    0.445987587	0.884042756
                    1.605074139	1.520288168
                    2.253731874	1.772979367
                    2.107138363	1.805787114
                    1.851496899	2.009453852
                    1.673462221	1.457891498
                    1.578530454	0.837830637
                    1.67697523	0.43146432
                    1.391175305	0.201934418
                    1.355433035	0.241458559
                    0.741217214	0.313228319
                    0.805736835	0.1109592
                    1.088755069	0.406739079
                    2.084948457	0.819716927
                    2.345079558	0.544349862
                    2.231116528	0.5747248
                    2.103747598	0.358502121
                    1.678264491	0.017482077
                    0.750516636	0.176414357
                    0.428897088	0.102972512
                    0.252079642	0.277153566
                    0.200638604	0.070445872
                    1.009462989	0.284754492
                    1.340605874	0.071725838
                    1.18842196	0.699855839
                    1.175329395	1.002621961
                    1.018457928	1.012160718
                    1.141246724	0.385410601
                    0.798635358	0.731942394
                    0.600018132	0.807583547
                    0.758299725	0.72265374
                    1.279910581	0.117245295
                    1.994472025	0.100630922
                    3.203247985	0.214741344
                    3.847297782	0.57300983
                    3.478036025	0.976006994
                    2.720084195	1.868113385
                    1.643073602	1.593161532
                    2.003773905	1.306230705
                    5.827261345	0.422601557
                    9.931272463	0.187867225
                    11.57802852	0.173435088
                    10.00455767	0.495297468
                    6.334754927	0.180385908
                    2.924908025	1.354765542
                    0.762916018	2.073288521
                    0.298304069	2.925511412
                    0.254853416	2.437349727
                    0.766277906	1.893309469
                    0.249323009	2.26234556
                    0.295244002	2.430867558
                    0.326896353	1.764142655
                    0.4292154	1.134102355
                    0.049844661	0.846556936
                    1.09559658	0.526741292
                    3.591341349	2.242993052
                    5.140008683	2.346930108
                    6.546402605	3.347002458
                    6.550170811	2.63926772
                    5.854348066	2.478442716
                    5.919006278	2.825049293
                    4.764863072	3.078429634
                    3.678484084	2.974671605
                    3.332657405	2.181461834
                    3.151287543	1.548612581
                    3.431356509	1.456109968
                    3.491403047	1.003709879
                    3.874134785	0.637612598
                    3.673223502	0.120267621
                    3.319755963	0.13021218
                    1.873104472	1.078656976
                    0.774730567	2.067511863
                    0.235373006	1.942204895
                    0.804818029	1.340160509
                    1.06922962	0.569065949
                    1.613965801	0.419268832
                    1.439403492	1.727551357
                    1.811028283	1.197022341
                    2.487667988	0.029199374
                    2.068919433	0.378159706
                    1.914682141	0.592645653
                    1.505253635	0.479471546
                    1.275999268	0.242108219
                    0.96847192	0.302868576
                    1.059371698	0.838206637
                    1.333299997	1.908495963
                    0.716763494	1.758812482
                    0.486405589	1.034126332
                    0.872183622	1.359504492
                    1.422905024	1.756085992
                    1.897266181	1.438201055
                    1.918901373	0.961417778
                    0.804618469	0.947026837
                    1.151393038	0.481802527
                    3.230269054	0.048596005
                    4.796821957	0.038212018
                    5.193044344	0.042202897
                    4.159266879	0.843237544
                    2.602544323	1.177368774
                    1.072903345	0.925158966
                    1.170192514	0.428953333
                    2.098456072	0.831780378
                    3.534648585	1.903633691
                    4.067352668	2.111461722
                    4.159881904	1.983863187
                    4.116197944	2.069833417
                    3.775583836	1.982667726
                    2.672169265	1.152776673
                    0.936092673	0.022801448
                    2.437409754	2.76711596
                    5.441006342	5.405459885
                    6.624030285	6.76105061
                    6.674705103	6.418847341
                    6.386493875	5.268974512
                    5.246984566	4.229946225
                    4.24299386	3.148381398
                    3.747335909	2.75941329
                    3.138687332	2.569480411
                    2.159204606	2.163161112
                    0.881153204	1.030524364
                    0.294664357	1.109443769
                    0.231798768	0.850066794
                    0.459856788	0.170644732
                    0.844408392	0.508095198
                    0.475503578	0.495147877
                    0.251358662	0.602390164
                    0.213749345	0.485422479
                    0.513702782	0.015376221
                    0.558900144	0.33931017
                    0.466754686	0.252527062
                    0.959363427	0.099681065
                    0.836366603	0.651044771
                    0.565264764	1.207474588
                    0.131353384	1.42778442
                    1.00350343	0.840941309
                    1.642403037	0.198677568
                    1.786366079	1.44837333
                    1.401892305	1.102261747
                    2.340458753	1.193619193
                    2.935227186	1.673066523
                    2.955848043	1.277404065
                    2.963280453	1.328510272
                    2.737167015	0.479909085
                    2.524912168	0.051021919
                    1.995650664	0.384528875
                    2.260862165	0.442882978
                    3.130472681	0.468004123
                    3.729723976	0.503488551
                    4.418587796	0.549924532
                    4.781708538	0.530565975
                    4.990948969	0.545550493
                    3.807333076	1.130139916
                    3.349894384	0.926400765
                    3.198352194	0.555202862
                    2.353358522	0.46835652
                    0.85782431	0.76220824
                    0.464773036	0.985467308
                    1.552935433	1.030311499
                    2.214550453	1.404658156
                    2.01053654	1.052057648
                    2.651422157	1.007427595
                    2.838951366	1.653459681
                    1.508243959	0.985394169
                    0.47456366	0.106528374
                    0.232235296	0.168484589
                    0.08248432	0.613612687
                    0.009444708	0.883213666
                    0.786085316	0.327392343
                    0.211033222	0.102595677
                    0.202534057	0.744185946
                    0.602733349	0.837788306
                    1.381355588	0.749514001
                    1.797362246	0.71930063
                    1.667436557	0.259765964
                    1.500050654	0.33042832
                    1.015956407	0.287828318
                    0.047780836	2.027803099
                    0.806190564	1.61492209
                    1.867717132	1.445572806
                    2.518801305	0.157342473
                    2.868021661	1.094033191
                    3.408246813	2.707479017
                    4.772579212	4.242903413
                    6.007360739	5.277370766
                    6.627959689	6.152587279
                    6.637861196	6.287148815
                    6.107661133	6.388479735
                    5.49106501	6.281357135
                    4.306646578	5.664879562
                    2.593746997	4.354360604
                    1.435495541	3.512664534
                    0.340281371	1.293658673
                    1.370006764	1.015866408
                    1.509291536	1.289321752
                    2.204091757	0.387409351
                    3.198539216	0.760157721
                    3.85662636	1.800792591
                    3.191738015	1.955813773
                    1.649317052	2.188725448
                    0.895346264	2.000728488
                    3.65436422	0.756705526
                    6.145738967	0.567347715
                    7.997511088	0.49062662
                    9.182175392	0.548174532
                    8.557730278	0.994755846
                    7.427377654	0.698561574
                    5.892979174	0.234664543
                    3.583445291	0.184814201
                    1.584620481	1.125523069
                    0.180246792	2.129380448
                    2.198738518	2.765200035
                    3.615245206	2.515445505
                    4.509983258	2.145611156
                    6.142128567	1.448619944
                    6.91618729	0.956653268
                    7.054490654	0.17016221
                    6.982489008	0.233570527
                    7.239382634	1.061279061
                    7.256675154	1.707151085
                    7.379663438	1.84179375
                    8.101149968	2.212941625
                    8.289167927	1.909330675
                    8.581003736	1.777542222
                    9.638655136	2.594751869
                    10.46294133	3.264124978
                    10.38822035	3.939992077
                    9.771757177	3.977828301
                    7.5498612	3.192166752
                    5.697015439	3.911591208
                    2.871365447	4.126811894
                    1.107086747	4.701132092
                    0.074120541	4.02359532
                    0.009950722	3.014750679
                    0.000112592	1.744174802
                    0.318893896	0.290475084
                    0.205676671	0.357086542
                    0.174410266	0.805617814
                    0.501336194	0.253977802
                    0.325627367	0.263843546
                    0.508192632	0.041341617
                    1.120857666	0.683789756
                    1.882587181	0.814065843
                    2.308547995	1.186936219
                    1.489641651	1.349444059
                    0.163957775	0.75262084
                    0.408705814	0.786486131
                    0.429227375	0.812928329
                    0.832255158	0.909985812
                    0.183711465	0.76325489
                    0.963609592	0.390567908
                    1.744277574	0.470615552
                    3.477692783	0.949934556
                    4.060773097	2.108260779
                    3.894549186	2.280870514
                    3.815193247	2.006701524
                    3.17811739	1.972593976
                    2.456323099	1.711096737
                    2.065389842	1.886625031
                    1.304320358	2.23079797
                    1.41350742	1.795713619
                    1.163969905	1.370197861
                    0.388207586	1.320157195
                    0.056396278	1.22490563
                    1.335243416	2.252220843
                    2.951032687	3.301200197
                    3.794866311	4.188570384
                    4.610627146	4.631759419
                    4.495070591	4.695415951
                    3.828285402	4.503951918
                    3.553107089	3.689431401
                    2.828184474	2.738096379
                    2.581103893	2.05926626
                    2.183883704	1.616383227
                    2.181482391	1.330857965
                    1.853553292	0.429649595
                    2.02944144	0.392685583
                    1.907194216	0.446379853
                    2.69042802	0.270308931
                    2.177098321	0.106361502
                    2.138652289	0.249571779
                    1.014301367	0.082980877
                    0.647891394	0.086331334
                    0.158182053	0.256710526
                    1.554570296	0.057307292
                    2.001300857	0.133226243
                    1.975891237	0.36235092
                    0.919569671	0.171330082
                    0.64737585	0.774339205
                    1.312705025	1.082398983
                    1.682788111	0.984471341
                    2.232676716	1.106244575
                    1.860504895	0.50743399
                    0.958216089	0.127052589
                    0.816138093	0.318471472
                    0.174035285	0.801671799
                    0.039823527	1.02161453
                    0.223925957	0.756042691
                    0.257137296	1.55949513
                    0.815995215	1.763752585
                    1.219743875	2.18195302
                    2.327378151	2.86378997
                    2.291053076	2.667745709
                    2.27587933	1.893098042
                    2.353806167	0.811466916
                    2.337097665	0.174340413
                    2.033907155	0.001007964
                    1.617645761	0.113091827
                    1.501755318	0.333482009
                    1.103866864	0.323327507
                    0.338889905	1.205504238
                    0.730156555	1.051993964
                    0.324130218	0.290030842
                    0.619935734	0.123781893
                    0.973703877	0.003773032
                    1.117364883	0.442252411
                    1.310111935	0.281940368
                    1.04811249	0.736173912
                    1.185563727	0.312722994
                    0.879291314	0.567792805
                    0.438850647	1.073466445
                    0.713404354	1.16787287
                    1.250584943	0.815934821
                    1.471017368	1.086401287
                    1.893926944	0.397099141
                    1.187916727	0.107543659
                    1.251794759	0.019427935
                    1.140250167	0.492445704
                    2.493263203	1.7527403
                    2.976765503	2.03324031
                    2.698518109	1.374309704
                    1.91689357	0.642273263
                    0.449760443	0.2251013
                    2.208352407	2.073931746
                    5.420706707	4.705974537
                    6.999059272	5.887687423
                    6.657547057	5.642865501
                    4.514971303	4.41719829
                    2.314381431	2.150163227
                    0.352556713	0.363247471
                    2.054157908	2.082234809
                    2.266251636	2.703479794
                    1.931657825	2.108394728
                    2.349189167	2.147995384
                    2.456614909	1.544257426
                    4.102396717	1.922190498
                    4.59851353	1.725099953
                    4.466942835	1.43497548
                    4.945370644	1.676152569
                    5.029269911	0.529466065
                    5.877094093	0.391763714
                    6.731261684	0.698195462
                    7.227319774	1.273189288
                    6.770495923	1.33990884
                    3.043136227	0.942174157
                    2.73938771	0.383557333
                    8.292748185	0.836957079
                    11.14146905	1.068222444
                    10.89998321	1.441405276
                    7.8477537	1.445704552
                    3.881459001	0.657318694
                    1.229319727	0.420536237
                    0.151802554	0.626719191
                    0.338959298	0.604557273
                    0.387559321	0.764651312
                    0.949238456	0.924904076
                    1.237537579	1.229288008
                    0.662989448	1.636152554
                    0.18909661	1.157425543
                    1.313745273	1.019692126
                    3.034620146	0.48391257
                    4.973469248	0.270863213
                    5.228949769	0.377641286
                    6.107111153	0.627870926
                    6.346711955	1.039368885
                    6.463928348	0.523363148
                    6.01842572	0.418946357
                    6.129306327	0.148292867
                    5.885206875	0.083117043
                    5.400314745	0.456158977
                    5.30741399	0.417008808
                    5.185567849	0.130103982
                    5.28741089	0.184842404
                    4.52806968	0.635955215
                    3.281614124	0.080289455
                    1.860780034	0.944928066
                    0.477285195	1.499027044
                    0.543129344	1.747703221
                    0.99117439	2.135305145
                    1.837544016	2.691718453
                    1.662992785	1.82573436
                    1.218159941	1.871047295
                    0.982940291	1.170679414
                    1.257340746	1.404289298
                    1.326750688	1.453139175
                    1.454256346	1.889748784
                    1.2142023	2.452457841
                    1.460993178	1.989325639
                    0.88523928	2.351076745
                    0.020536495	2.46094073
                    1.298429447	1.565371342
                    1.593141325	0.955052475
                    1.496312322	0.588479179
                    1.851174954	0.20093222
                    2.085676214	0.159066356
                    2.62424691	0.799110688
                    2.546163145	1.444870299
                    1.407583744	0.893425047
                    0.285995392	0.273232311
                    0.747686355	0.808370734
                    1.024918712	0.421646425
                    0.738671823	0.118141849
                    1.989034557	0.441473764
                    1.702245246	0.179613084
                    1.166379864	0.204626137
                    2.306971683	0.061558852
                    2.57582779	0.44290718
                    3.586969945	0.681468861
                    4.753189266	1.21488824
                    4.961008297	0.855205079
                    5.006438611	0.561777316
                    4.352154622	0.04895993
                    3.768548534	0.913892908
                    2.579616518	1.012170369
                    2.121891924	0.719490878
                    2.028822771	0.355906069
                    1.786621049	0.266837288
                    0.561963739	0.056599276
                    0.242273556	0.543912841
                    0.129143488	1.010553136
                    0.791473724	1.089897496
                    2.390515834	0.641095194
                    5.261407031	0.34230754
                    8.199265051	1.192464408
                    8.761935179	1.275761547
                    7.270051346	1.030510219
                    4.012524679	0.374642241
                    1.131312817	0.607738023
                    0.565935203	0.572398514
                    0.894597967	0.446804002
                    2.114236911	0.77717157
                    0.898790677	0.741924739
                    0.571649216	0.33582736
                    1.867492608	1.074686699
                    2.006170278	1.755010517
                    3.309613688	2.810988705
                    3.207290665	1.705721536
                    2.541655977	2.354900935
                    1.259676411	1.985514626
                    0.351516841	1.797157962
                    1.725351935	0.808933573
                    3.06997675	0.008000337
                    3.650762218	0.147391982
                    4.109379503	0.812154022
                    3.093117852	0.599820332
                    3.12374343	0.031647907
                    2.866744581	0.728819927
                    2.999900732	1.146777569
                    1.825016107	0.565761646
                    1.544000652	0.864617464
                    0.402893542	0.409623615
                    0.640158593	0.178344346
                    0.528593957	0.08133832
                    0.67784934	0.085475274
                    0.028039029	0.307673009
                    0.301876636	0.44800089
                    0.486606367	0.042755444
                    0.425373919	0.420769718
                    0.109502549	0.412271189
                    0.069888042	0.757729606
                    0.006600146	0.683817906
                    0.131694576	0.550592591
                    0.351097395	0.586714649
                    0.220137251	0.516675072
                    1.064988687	0.978874871
                    0.599680947	0.132526511
                    0.011826401	0.532494594
                    0.269765088	0.888358936
                    0.631024259	1.081639957
                    0.193166122	2.754409469
                    0.124928552	3.180317891
                    0.565826515	3.139042497
                    1.443815819	2.368867937
                    1.08034962	2.571789083
                    0.365979997	2.685695299
                    0.404606464	2.065219198
                    1.224829694	1.555503007
                    2.385475213	1.745718628
                    2.636133661	0.778941249
                    3.295216553	0.47848641
                    3.087876425	0.001888955
                    2.903598319	0.624126003
                    3.289097643	0.887539806
                    3.061268744	0.383709503
                    2.283096818	0.438664682
                    1.372209602	0.531096105
                    0.562073138	2.1295723
                    2.620481702	1.743628007
                    3.824789567	2.448993764
                    4.500157544	2.209654682
                    4.975960173	2.005939097
                    5.686726454	1.935470052
                    6.986668943	1.759032192
                    7.902919796	0.89995796
                    8.54953357	0.177564531
                    8.069885683	0.770001703
                    6.828346751	1.027662064
                    6.164326697	0.678340624
                    5.55433258	0.030091479
                    3.611562243	0.730559421
                    2.840103565	0.934555953
                    1.882882689	1.617483
                    1.560211092	1.50550031
                    1.714881434	0.815291941
                    1.442061118	0.168571859
                    1.522220911	1.008948153
                    0.947009726	1.268935969
                    0.385975895	1.685012007
                    0.723756226	1.558603916
                    1.084131685	1.250015796
                    1.928394153	0.638802465
                    2.983370943	0.41424336
                    3.190428642	0.039423426
                    3.444913973	0.468543923
                    3.65306026	0.789019968
                    3.16085674	0.622000348
                    1.736131483	0.066913264
                    1.216511334	0.049098755
                    0.153688368	0.842628546
                    0.325581782	0.275071414
                    0.924003676	0.337628492
                    1.315895189	0.809895036
                    0.601575281	0.159880129
                    0.592140267	0.546436723
                    0.026183894	1.076343435
                    0.891051719	0.736706594
                    1.648932653	0.719064674
                    1.694084698	1.474063041
                    0.470945019	0.92411888
                    1.06564145	0.739297085
                    3.097935794	1.359157148
                    4.219260442	1.84799106
                    4.521259841	1.122413459
                    4.5607552	0.328273928
                    4.434768888	0.153394186
                    4.203551565	0.516489521
                    4.459621174	0.452678799
                    4.595925024	0.741307448
                    4.852880574	1.511887959
                    5.360203923	1.805550928
                    4.910173073	2.442389793
                    4.74385199	1.758301499
                    3.791377964	1.246405483
                    1.583054308	0.313050905
                    0.996225407	0.577700686
                    1.465744586	1.132843198
                    2.312559127	0.829808931
                    2.550845433	0.777726214
                    1.87892573	1.56060137
                    1.579199935	1.623783398
                    2.585528816	1.615226348
                    4.224575644	1.358669358
                    5.516843411	1.493411567
                    6.614988297	2.414520821
                    6.407462619	1.578748641
                    4.432861999	0.601429023
                    2.550212353	0.4355539
                    1.119274834	0.533526364
                    0.331646041	0.483266441
                    0.500124033	0.303314469
                    1.828510591	0.683168719
                    2.919550161	0.312229209
                    3.864670203	0.614438551
                    4.759639016	1.09795179
                    4.227547554	1.519869045
                    3.254708178	2.208242797
                    2.597473437	1.865355136
                    2.287198302	1.542721848
                    1.211351723	0.47470655
                    0.538000304	0.016520173
                    1.171158572	0.871188895
                    1.466391498	2.145371011
                    0.292670699	3.152487177
                    0.595332308	2.873651647
                    0.279620634	2.412999672
                    1.541161766	2.472050728
                    2.457178256	3.133658914
                    2.633079787	2.645541295
                    2.268412393	1.61199491
                    1.78190437	1.063957886
                    1.485872739	0.856858876
                    1.332019875	1.263376102
                    1.05030154	1.541568597
                    2.012465494	2.584008937
                    2.178464537	2.612715638
                    1.997531172	1.857957018
                    2.532692588	2.532852728
                    2.915883518	2.730484941
                    3.147487117	3.164740713
                    3.114833554	3.056683329
                    2.830367067	2.406975236
                    1.672376246	1.425525993
                    1.233104986	1.191861446
                    1.273005288	1.058482061
                    0.628197184	0.49473656
                    0.634615309	0.041570334
                    0.506718185	1.135564634
                    1.081807061	1.985761102
                    1.3574345	2.464690304
                    2.268643254	2.75299391
                    3.358805148	3.189408384
                    3.327029066	2.796324046
                    2.911348639	1.908943048
                    2.807508632	1.598137153
                    2.36099198	1.13044072
                    2.189483357	0.987841031
                    2.235258152	1.698418677
                    1.753884185	1.470382068
                    1.033081799	1.000390587
                    1.378195404	1.222891194
                    2.269500449	0.935475779
                    2.462001488	0.312253225
                    2.344155557	0.042553623
                    3.81047264	0.423386357
                    4.459598038	1.110518801
                    4.846754324	2.088220025
                    5.404171228	2.040855521
                    6.370692568	2.315196039
                    6.663244213	1.525571031
                    5.68588199	0.503197808
                    3.197255766	2.960507232
                    1.317353388	5.339420263
                    0.978016838	6.733580426
                    1.422488129	5.43799586
                    2.710097743	4.033883607
                    2.577745065	2.497360054
                    2.683206732	2.181683615
                    2.435542226	2.168153413
                    1.825809838	1.959397329
                    0.410775735	0.614717455
                    0.255029503	1.061946416
                    1.109728819	1.893160161
                    1.413427834	2.480132978
                    0.476524137	2.266646567
                    0.382641589	1.914463441
                    0.081903125	0.852756241
                    1.087025249	0.788017377
                    4.164891857	0.824899784
                    8.077215247	0.822872025
                    11.43198387	0.513896425
                    12.66531767	0.245989411
                    12.62508668	0.287392813
                    11.85094689	0.500584932
                    10.17096469	0.609741148
                    7.334030232	0.288456141
                    4.355305573	0.560494578
                    2.074584555	0.532301847
                    0.035529266	0.605828283
                    1.481850333	0.278234908
                    3.01276506	0.072002651
                    4.322760342	0.071678195
                    4.249815944	1.257276587
                    4.293619575	2.909405124
                    4.192281219	2.158883758
                    3.245166234	1.73347783
                    2.550024815	1.877745496
                    3.418264048	0.591752256
                    2.402890057	0.962455104
                    2.563882268	0.522250133
                    2.598006244	0.460670833
                    1.196547	0.717770814
                    0.292010352	0.394675046
                    0.117427389	0.03461955
                    0.52272685	0.3367439
                    0.880994164	1.012847559
                    0.706870616	0.406413619
                    0.745730632	0.577757169
                    0.92522655	1.174666339
                    0.203641443	1.197968424
                    2.283191652	0.057439443
                    4.572675214	1.005506093
                    4.64896313	1.008409835
                    4.190397521	0.820583024
                    2.369083118	0.042189055
                    0.561345226	1.69067345
                    0.624541402	2.180065847
                    1.081575697	2.47973652
                    1.362271498	2.90941772
                    1.12378338	2.428620058
                    0.873379931	1.945680152
                    0.476104673	1.690594774
                    0.475228219	1.257142784
                    1.144823322	0.033573887
                    0.815113194	1.471289759
                    0.087616412	1.433876423
                    0.219185813	0.87674933
                    1.345196192	1.323693814
                    1.256860242	0.667233097
                    1.574390429	1.424713938
                    1.273947813	1.364205861
                    1.670693995	1.136995642
                    1.990176248	1.139174325
                    2.607490224	0.662773022
                    1.694825421	0.397784607
                    0.289551413	1.306999829
                    0.141908722	1.583838825
                    0.274331844	0.785443049
                    0.391396051	0.661148028
                    1.081336307	0.419844286
                    2.355527329	0.196107077
                    3.035294037	1.009085031
                    1.999881275	0.516599765
                    0.441839553	0.188564413
                    0.061098614	0.439838582
                    1.0493012	1.28281307
                    1.745548061	1.780608553
                    1.536748448	2.385049141
                    1.743496324	3.139354427
                    1.011448642	3.327086878
                    0.140237972	1.93396383
                    0.669546319	0.240956677
                    1.103269829	0.71725782
                    0.655649971	1.003599607
                    0.208355087	1.128630822
                    1.064763341	1.598648512
                    1.317621758	1.562227573
                    1.811914805	1.824151665
                    1.504274475	1.518838307
                    0.65295162	1.758590675
                    1.363922468	2.505794076
                    1.52674239	1.457278903
                    0.932703714	1.18931158
                    0.337220554	1.092874776
                    0.030721915	0.982183918
                    0.050274595	0.294139685
                    0.132413181	0.364762005
                    0.302109531	0.108502804
                    1.597409225	0.41329009
                    2.196826565	0.260769176
                    3.214840997	0.171845521
                    3.723313561	1.279122823
                    3.044809025	2.119618232
                    1.260182264	2.341037284
                    0.634414682	1.70988021
                    0.719872926	0.572580259
                    1.340090847	2.225312858
                    1.559537363	2.477878942
                    2.021518494	2.413786057
                    2.704761512	2.336447893
                    2.759259304	2.103442602
                    3.548915207	2.691086153
                    3.623098939	2.369267687
                    3.758953558	2.605810529
                    3.323467159	1.21707232
                    2.822868556	0.346319384
                    2.143802939	1.734849609
                    0.990199398	2.433240376
                    0.414646463	2.251094111
                    0.349235196	1.921244531
                    0.405220629	1.895016525
                    0.035102067	1.736765876
                    0.303841928	1.675802597
                    1.172600707	1.841166727
                    1.559458836	1.750589981
                    1.536634495	2.153249486
                    1.946697689	1.455244268
                    1.018486545	2.106036294
                    0.946840796	1.186445186
                    0.91385879	0.750204421
                    2.320860885	0.492823106
                    4.195722045	1.530812355
                    6.385917943	1.104069658
                    5.543858499	0.45554191
                    2.830107235	0.041463247
                    2.019589974	0.168025113
                    5.685874655	0.865571715
                    6.171263057	1.020483115
                    4.28541813	1.317434235
                    1.176971787	0.613380565
                    2.493023323	0.413318214
                    5.416367572	1.575628047
                    7.244102253	2.143826315
                    7.468406267	2.67372915
                    6.67211847	2.970318658
                    5.382699562	3.010520337
                    3.620052079	2.853955404
                    1.806825821	2.353405676
                    0.533180877	1.399811751
                    1.601591825	0.218622011
                    2.696245966	0.175760134
                    3.297862995	0.137547081
                    3.10739295	0.06316084
                    3.388512072	0.175653021
                    3.530306119	0.363424805
                    3.216873385	0.465085762
                    2.790909968	0.373377246
                    2.935069779	0.167803674
                    3.601695884	1.001960335
                    3.3529797	1.178044619
                    3.861465048	1.213929812
                    4.233683509	2.557410672
                    3.533824825	2.997694439
                    0.873956053	1.837936692
                    1.305872814	1.713063687
                    2.563391215	0.991028681
                    2.526187071	0.525263137
                    2.438643823	0.540513525
                    2.213089218	0.42287822
                    2.413969752	0.12225661
                    2.281036034	0.092367188
                    2.625491172	0.018678852
                    2.897461285	0.081067458
                    2.910634877	0.274020346
                    2.520862589	0.063943081
                    2.021608645	0.967698725
                    0.639029022	1.176330082
                    0.285802877	1.675877153
                    0.226795256	1.325288425
                    1.021704487	0.440126858
                    2.382929	0.50894936
                    3.785263759	0.783225397
                    5.095703179	0.490867435
                    5.477215823	0.526524194
                    5.200238455	1.104028034
                    4.688090595	0.331960156
                    2.330313499	0.560604893
                    0.467398285	0.060044195
                    1.838940079	0.684031431
                    2.067292791	0.751874245
                    1.47456958	0.94350541
                    0.228388305	1.256678988
                    0.475123523	1.785384269
                    1.044387145	2.470263387
                    1.288820378	2.862658144
                    0.639328724	2.628958944
                    0.633272143	2.281916465
                    0.043161811	1.599312554
                    1.376841024	1.041118655
                    2.310379566	0.579920084
                    3.383056601	0.109071717
                    3.639718098	0.003220168
                    3.156049883	1.177018169
                    3.062237254	1.007930781
                    2.761885005	0.999001007
                    2.731874794	1.459324708
                    2.526708708	1.28880218
                    2.662193198	1.407627864
                    2.961006985	1.12552701
                    2.963657924	0.21779417
                    1.532283355	0.938224186
                    1.744351151	0.228282117
                    1.400687308	0.01382232
                    1.204288534	0.260468691
                    1.090918584	0.750209422
                    0.198819046	0.526376891
                    0.588213677	0.976282531
                    0.218277967	1.59244163
                    0.890116405	2.19618109
                    0.739879266	2.497349095
                    0.012159494	1.440741449
                    0.53710727	0.341590886
                    1.656729528	0.769830458
                    2.739844768	2.037139217
                    5.319788258	4.805086723
                    6.614378346	6.372932932
                    7.56633232	7.268361995
                    8.718180904	8.256574325
                    7.617007839	6.70913785
                    7.095910967	5.787754246
                    4.548374898	3.764921823
                    1.151266308	1.345903053
                    2.648551262	1.692808494
                    5.798024972	3.622734834
                    8.202900393	5.040608577
                    9.656136249	5.027584208
                    10.49197757	4.889154911
                    10.24546918	3.712272549
                    7.851741912	1.113833716
                    3.833175968	1.80249598
                    0.067472814	3.371027496
                    2.865904652	4.601632115
                    3.342375074	3.69838716
                    3.193610431	2.097821029
                    0.923240383	0.682192453
                    1.532878476	3.406596037
                    3.820796462	5.163020598
                    5.904306585	6.409709814
                    7.475878657	7.158126608
                    8.626271166	7.624936329
                    8.955184841	7.012905187
                    8.049626727	5.903402345
                    7.812064563	5.168760822
                    7.429545045	4.082262918
                    6.148909395	2.721934206
                    5.535831051	2.028641699
                    3.975679254	1.326524487
                    2.336500479	1.236055481
                    0.581616845	1.370204616
                    1.010625801	0.910605051
                    0.231681987	1.277330491
                    0.139973885	0.897143206
                    0.741502243	0.192280161
                    0.714445956	0.254665529
                    1.006266001	0.384205921
                    2.192204636	0.005793268
                    3.350438907	0.118672555
                    3.79680246	0.664495549
                    3.831227187	0.727985174
                    3.593117827	0.154603473
                    4.241709417	0.298093869
                    5.22590151	1.47793016
                    6.198233779	1.643883725
                    5.737768321	1.768658117
                    4.888681403	2.115553712
                    4.860213536	0.860675074
                    5.434734533	0.185669651
                    4.451234518	0.208897998
                    3.811530964	1.100705723
                    3.397187811	1.433112538
                    2.139359686	0.81938806
                    2.124276733	0.713240071
                    0.143987502	0.232758103
                    1.378725263	0.666306764
                    1.923205198	0.598427077
                    2.160542615	0.907754469
                    1.076242414	1.849586437
                    0.485138302	1.432054166
                    0.040047544	0.172713644
                    1.498320055	0.041450408
                    2.802356729	0.462008343
                    3.604642296	0.764138621
                    2.679266247	0.04969392
                    1.763918644	0.559438439
                    0.663746814	0.386752872
                    0.029652309	0.017883068
                    0.475623808	0.428787014
                    0.715981999	0.807989053
                    0.847086618	1.530713246
                    1.113325002	0.933087711
                    0.900817553	0.394840198
                    0.762556609	0.235907392
                    0.365211935	0.056362999
                    0.026402081	0.075383009
                    0.446072951	0.341354494
                    1.286759436	0.920623372
                    2.608820593	1.727231957
                    2.968930234	1.799508118
                    2.191146909	1.066462558
                    2.766181319	1.210910158
                    2.954821887	2.08678665
                    3.340967105	2.47395951
                    3.353892376	2.85966433
                    2.893336369	2.578796865
                    3.364869818	2.675625437
                    3.390584576	2.325747036
                    3.617156083	1.8164257
                    3.418605033	1.007530489
                    3.569211256	0.190313997
                    4.22805716	0.698398876
                    4.283175458	0.768024064
                    5.110632269	1.343661023
                    5.935890002	1.811441833
                    6.466966531	1.582911665
                    8.36023962	2.397647071
                    8.916307735	3.047365038
                    8.238962565	3.190980362
                    6.803892133	2.361544803
                    3.292325061	1.493365929
                    0.040737217	1.544794903
                    2.210439856	0.451503258
                    2.742841126	0.437825839
                    2.676088821	0.253145945
                    1.7897102	0.961571716
                    0.977013096	1.146233945
                    0.083743903	1.296800795
                    1.893101777	2.011597447
                    3.619161521	2.876581918
                    4.57840384	2.985783462
                    5.647010152	3.734924753
                    6.780269703	4.23135935
                    6.756053545	4.109362769
                    6.634121134	3.545373218
                    7.318170146	3.231425532
                    7.108268027	1.778653865
                    7.563798577	0.828631511
                    6.756108026	0.302210646
                    5.21289158	0.381475131
                    2.337478743	1.123614296
                    1.815273757	1.887634494
                    7.979247533	2.181480444
                    10.59395527	0.936773226
                    11.40290129	0.468241355
                    9.4763129	0.221160691
                    5.234003942	0.895719642
                    1.662620305	0.329927102
                    0.177403942	1.878045047
                    1.126797481	4.006586288
                    2.565563257	4.055164805
                    2.401271906	2.560525461
                    2.152055411	0.904118288
                    0.744118676	0.181065972
                };
            \addplot[color=red,thick] table {
                    -0.8 -0.8
                    15 15
                };
        \end{axis}
    \end{tikzpicture}
    \vspace*{-9pt}
    \caption{
    The comparison of error between $\big|\estg_{[5]}-{\bmx_t^*}_{[5]}\big|$ and $\big|\estg_{[5]}-{\bmx_t^*}_{[5]}\big|$ for $300 < \tau_t \leq 350$.
    }
    \label{fig:err5}
\end{figure}